\documentclass[10pt, a4paper]{article}
\usepackage{stfloats}

\usepackage{lrec-coling2024} 
\usepackage{lrec-coling2024} 

\title{Toward Automatic Filling of Case Report Forms: A Case Study on Data from an Italian Emergency Department}

\name{
\begin{tabular}{c}
Gabriela Anna Kaczmarek\textsuperscript{1},
Pietro Ferrazzi\textsuperscript{1,2},
Lorenzo Porta\textsuperscript{3}, \\
Vicky Rubini\textsuperscript{4},
Bernardo Magnini\textsuperscript{1}
\end{tabular}
}

\address{\textsuperscript{1}Fondazione Bruno Kessler, Povo, Trento, Italy \\ \textsuperscript{2}University of Padova, Padova, Italy\\ \textsuperscript{3}Emergency Medicine Unit, Fatebenefratelli Hospital, Milan, Italy\\ \textsuperscript{4} University of Milan, Milan, Italy\\ \{gkaczmarek, pferrazzi\}@fbk.eu\\}

\abstract{
Abstract: Case Report Forms (CRFs) collect data about patients and are at the core of well-established practices to conduct research in clinical settings. With the recent progress of language technologies, there is an increasing interest in automatic CRF-filling from clinical notes, mostly based on the use of Large Language Models (LLMs). However, there is a general scarcity of annotated CRF data, both for training and testing LLMs, which limits the progress on this task. As a step in the direction of providing such data, we present a new dataset of clinical notes from an Italian Emergency Department annotated with respect to a pre-defined CRF containing 134 items to be filled. We provide an analysis of the data, define the CRF-filling task and metric for its evaluation, and report on pilot experiments where we use an open-source state-of-the-art LLM to automatically execute the task. Results of the case-study show that (i) CRF-filling from real clinical notes in Italian can be approached in a zero-shot setting; (ii) LLMs' results are affected by biases (e.g., a cautious behaviour favours "unknown" answers), which need to be corrected.
 \\ \newline  }

\begin{document}

\maketitleabstract

\section{Introduction}

\begin{figure*}[t]
\centering
\includegraphics[width=\textwidth]{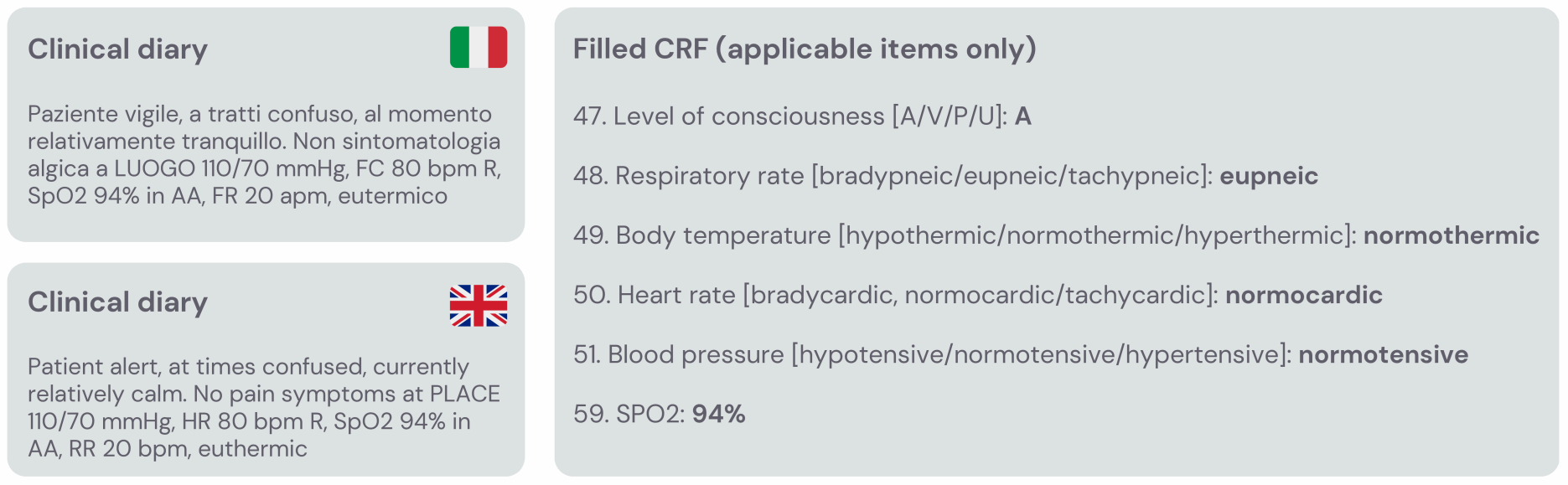}
\caption{Example of a clinical note, its English translation (provided only for reader comprehension), and the corresponding CRF items with assigned values. All other CRF fields default to \textit{unknown}.}
\label{f1}
\end{figure*}

Case Report Forms (CRFs) are a cornerstone of clinical research, providing a structured means of collecting patient data across diverse studies and healthcare environments. Traditionally paper-based, CRFs have long served as the primary instruments for recording patient information in a standardised format that supports consistency, comparability, and scientific rigour. Their role extends beyond simple data collection: well-designed CRFs ensure that study outcomes are interpretable, reproducible, and aligned with the specific research objectives they seek to address. These characteristics have been categorized in terms of 
interoperability \citep{Petavy2019},
clearness and identifiability of the used terminology \citep{Rinaldi2025}, 
consistency across different clinical settings \citep{Richesson2011},
applicability in different scenarios \citep{Bellary2014}
preserving the ability to target the specific context they are designed for \citet{LIN201549}.
Taken together, these perspectives highlight why CRFs are indispensable in modern clinical research. They anchor the collection of high-quality patient data, playing a central role in enabling interoperability and reproducibility.

The transition from paper-based to electronic Case Report Forms represents a significant milestone in the modernisation of clinical research. Electronic systems improve usability, reduce transcription errors, and facilitate direct integration with electronic health records \citep{Fleischmann2017}. The natural evolution of this transition is the automatic population of CRFs from unstructured clinical narratives. Early approaches, such as those described by \citet{MacKenzie2016}, explored the extraction of structured variables from narrative notes, while more recent contributions by \citet{Gutierrez-Sacristan2024} have expanded these methods. Despite the novelty of such frameworks, most solutions remain limited, relying heavily on keyword matching and vocabulary lookups. Such strategies often fail to capture the contextual richness of clinical language, leaving the advanced capabilities of modern Natural Language Processing (NLP), such as Large Language Models (LLMs), underexploited. Furthermore, the lack of publicly available datasets has stifled potential research in the field. 

Advancements in the integration of modern NLP techniques to Case Report Forms are presented by \citet{ferrazzi-etal-2025-converting}, which attempts to both fill the gap represented by the lack of data and to test LLMs to automatically fill CRFs from clinical narratives. Relying on semi-automatic data generation from scientific documents and medical exams, the authors do not tackle the lack of data coming from realistic settings. 
Filling this gap, our work focuses on data directly sourced from clinical settings. Here, we present an Italian dataset of pairs of clinical notes and filled Case Report Forms generated by the Emergency Department of the San Giovanni Bosco Hospital (Turin, Italy). An example of such dataset is reported in Figure~\ref{f1}.
This is, to the best of our knowledge, the first public dataset available for the CRF filling task on data coming from hospitals.
Our contribution can be summarised as follows:
\begin{itemize}
    \item we introduce a new annotated resource based on data from an Emergency Department in Italy, to be used for testing automatic CRF-filling;
    \item we showcase a pilot study on how Large Language Models perform on our data.
\end{itemize}

The dataset presented in the paper is made publicly available with Creative Common licence\footnote{\url{https://huggingface.co/datasets/NLP-FBK/SGB-crf-italian}}.

\section{Data Analysis and Task Definition}

In this section, we present the origin and structure of the CRF-filling dataset, the exploratory analysis conducted to characterise its properties, and the preparation steps required for its release. We then describe the Case Report Form (CRF) schema, the annotation process, and finally define the CRF-filling task together with the evaluation metrics used to assess model outputs.

\subsection{Dataset}

\paragraph{Clinical Notes.}
The dataset comprises 290 emergency-care notes collected from the San Giovanni Bosco Hospital in Turin, covering adult patients presenting between 1 January 2021 and 31 December 2023. Data access and use were governed by institutional protocols for the development and evaluation of NLP methods, under the eCreamEU project. The notes are distributed across seven categories, with approximately 30–50 notes in each category: anamnesis reports, triage records, nursing care notes, specialist consultation notes, medical visit reports, diagnostic reports, and discharge summaries. In the current release, the protocol required strict anonymisation: all identifiers, such as names, phone numbers, addresses and ID codes were removed by a third-party project collaborator. No pseudonyms or internal patient codes were retained either - consequently, notes referring to the same patient among different categories cannot be mapped together (for example, an anamnesis report, a nursing care note, and a discharge summary belonging to one individual cannot be connected). While this restriction limits longitudinal reconstruction at the patient level, it guarantees full compliance with privacy requirements. Future dataset releases are planned to refine anonymisation procedures so that cross-note linkage can be supported without compromising confidentiality.

\paragraph{Case Report Form.}
The Case Report Form (CRF) is composed of a structured list of medical items that need to be filled with the patient's information. In this study, the CRF contains 134 items designed to capture clinical information in the setting of an Emergency Department. These items are topically divided into seven groups: History taking, Clinical examination, Diagnostic test results, Laboratory test results, Imaging test results, Treatment, and Final diagnosis. Each CRF item is associated with a closed set of allowed values, tailored to the type of information it encodes. The most frequent value sets are: (i) binary categorical values (e.g., yes/no, positive/negative, present/absent), (ii) ordinal values, often used to indicate the normality or abnormality of a measurement (e.g., below norm / within norm / above norm for body temperature), and (iii) a measured/unknown distinction, used for tests that do not have a fixed categorical outcome and would otherwise require contextual interpretation. Importantly, for every CRF item, the value unknown is always a valid answer, ensuring that the schema accounts for information that is absent from the clinical note.

The CRF template was developed in collaboration with clinicians to ensure medical validity and practical applicability. Its design reflects the requirements of a documented research use case focused on dyspnea and transient loss of consciousness (Use Case 1 of the eCream project). 
\paragraph{Annotation Process.}
Annotation was performed by clinicians using the Label Studio software. Each entry records the character offsets of the supporting text in the note, the corresponding text fragment, and the CRF item together with its assigned value. 
\paragraph{Exploratory analysis outcomes.}
On average, each note contains 5.7 annotated items, though distribution is uneven: some are densely annotated, while about 20\% (58/290) are not filled at all. Label density generally remains modest, rarely exceeding 15 items per note, with a few cases reaching up to 40. Given that the CRF comprises 134 fields, this means that for any given note, well over 120 items on average are assigned the value unknown. Notes also differ strongly in length, with a mean of 131 words, with outliers reaching up to 400. This varies across categories, e.g. triage and discharge records are often shorter than the rest.

Coverage of the CRF is incomplete: roughly three-quarters of the items are represented (which means they were used as an annotation at least once), while the remaining quarter never appear in the data. Annotations frequently include redundancies (28\% of notes contain repeated annotations of the same item with the same value) and occasional conflicts (about 0.8\% of notes have the same item assigned to them more than once, but with a different value). These are not annotation errors but rather reflect how clinical concepts are documented over time, for example, when a patient’s condition changes during their stay. Such phenomena are expected to become even more prominent once future dataset releases enable linking multiple notes for the same patient.

At the group level, the distribution of annotations is highly uneven. Clinical examination (24 items, 676 labels) and history taking (46 items, 661 labels) dominate the dataset, together accounting for over 80\% of all observed annotations (1337/1648). By contrast, laboratory results (24 items, 148 labels) and imaging results (10 items, 85 labels) are far less represented, and the remaining groups—treatment (7 items, 29 labels), final diagnosis (20 items, 27 labels), and diagnostic test results (3 items, 22 labels) contribute only marginally. The CRF groups are not of equal size, but the observed skew exceeds what can be explained by group length alone. This reflects the clinical reality of the Emergency Department, where vital signs and history are systematically recorded for nearly all patients, while detailed test results, treatment actions, or final diagnoses are documented only in a subset of cases. 

Lastly, there are correlations between item groups and note categories, which are easily explained by the note's function: for instance, anamnesis notes align most closely with history items and discharge notes with diagnosis items, with similar patterns appearing for other category–group pairs as well.

\subsection{CRF-filling: Task Definition}
The task is defined as the automatic population of the 134-item CRF from each emergency-care note. For every item, a system must output one filling value from the item’s predefined set of allowed values. Notes are provided in Italian, while CRF items and allowed values are presented in English, requiring cross-lingual comprehension. Whenever the note does not provide sufficient evidence, the correct output is "unknown", which is always considered a valid answer.

Figure~\ref{f1} illustrates how a clinical note is mapped to CRF fields. The original Italian text is shown, while the right panel lists the CRF items whose values can be derived from the note. All other fields in the CRF default to unknown. An English translation of the Italian text is presented for clarity, but it is not included by any means in the dataset itself.

\paragraph{Evaluation Metrics}
Model predictions are evaluated by exact match against the ground truth value for each CRF item. The primary metric is micro-F1, which in this single-label setting is equivalent to overall accuracy and reflects how many item values are filled correctly across the dataset. Because the distribution of values is highly imbalanced—many fields are legitimately assigned unknown—we also report macro-F1, which averages performance across all possible values of a given item and thus gives a more balanced view of how well models recover less frequent but clinically informative outcomes. 

\section{Experimental Setting}

This section outlines the experimental framework within which the released dataset can be used for automatic CRF-filling. We describe the prompting strategy, model setup, and hyperparameters defining the experiment setting. Lastly, we report results from a pilot experiment conducted with LLaMA-3.1-8B Instruct \citep{grattafiori2024llama3herdmodels}  to illustrate feasibility.

\paragraph{Prompting.}
Experiments are designed in a zero-shot setting. Each CRF item is queried independently, with prompts structured in four parts: (i) a system role description, (ii) a brief introduction to the item group, (iii) an item-specific instruction including the list of allowed values, and (iv) the full clinical note in Italian. Predictions are made at the level of (note, item) pairs, so that the same note is processed independently for each CRF item, avoiding cascading errors across fields.

\paragraph{Model.}
The framework allows for the evaluation of a wide range of open and closed large language models in future studies. As an initial step, we conducted a pilot experiment with LLaMA-3-8B Instruct, chosen for its accessibility and suitability for zero-shot prompting on resource-constrained infrastructure. The aim of this pilot was not to optimise performance, but to validate that the dataset can be directly used for CRF-filling tasks.

\paragraph{Hyperparameters.}
Pilot inference was run using vLLM on a single NVIDIA A40 GPU (48 GB VRAM). The maximum context length was set to 4096 tokens, with temperature = 0 for deterministic outputs. Each clinical note required 134 separate model calls, one per CRF item.

\paragraph{Baseline.}
To contextualise model performance, we define a simple most-frequent-value baseline, where each CRF item is always filled with its most common value. For all items this corresponds to the label unknown.

\section{Results and Discussion} 

In table \ref{t1}, we report results from a pilot experiment using LLaMA-3 8B Instruct on the CRF-filling task, alongside a most-frequent-value baseline. Evaluation focuses on micro- and macro-averaged metrics to capture both overall correctness and performance on less frequent values.
\begin{table}[h]
\renewcommand{\arraystretch}{1.3} 
\centering
\begin{tabular}{|l|c|c|}
\hline
\textbf{Model / Baseline} & \textbf{Micro F1} & \textbf{Macro F1} \\
\hline
Most-frequent baseline & 0.963 & 0.353 \\
\textbf{LLaMA-3 8B (pilot)} & \textbf{0.920} & \textbf{0.548} \\
\hline
\end{tabular}
\caption{Overall performance of baseline and LLaMA-3 8B.}
\label{t1}
\end{table}

The most-frequent baseline achieves a deceptively high micro-F1 of 0.963, which stems almost entirely from always predicting \textit{unknown}. Its macro-F1 of 0.353 highlights that it fails to recover informative but less frequent values. By contrast, LLaMA-3 8B obtains a marginally lower micro-F1 (0.920) but improves substantially on macro-F1 (0.548), demonstrating that it attempts a broader range of predictions beyond the trivial baseline.

\begin{table}[h]
\renewcommand{\arraystretch}{1.3} 
\centering
\begin{tabular}{|l|c|c|}
\hline
\textbf{Ground truth value:} & \textbf{\textit{unknown}} & \textbf{\textit{other}} \\
\hline
Correct predictions (\%)    & 90.42 & 1.57 \\
Incorrect predictions (\%)  & 5.96  & 2.05 \\
Total share in dataset (\%) & 96.38 & 3.62 \\
\hline
\end{tabular}
\caption{Accuracy breakdown for LLaMA-3 8B predictions when the ground truth is \textit{unknown} or not.}
\label{t2}
\end{table}

The accuracy breakdown shown in Table~\ref{t2} confirms that the overall score is dominated by the majority class: 90.42\% of all predictions correspond to correctly identifying \textit{unknown}. Items with a non-\textit{unknown} gold label are rarely recovered (1.57\%).

\begin{table}[h]
\renewcommand{\arraystretch}{1.3} 
\centering
\begin{tabular}{|l|c|c|}
\hline
\textbf{Model} & \textbf{Precision} & \textbf{Recall} \\
\hline
LLaMA-3 8B & 0.562 & 0.569 \\
\hline
\end{tabular}
\caption{Additional macro-averaged precision and recall metrics for LLaMA-3 8B.}
\label{t3}
\end{table}

Precision and recall scores reported in Table~\ref{t3} are relatively balanced (0.562 vs.\ 0.569), suggesting that the model does not strongly favour precision over recall or vice versa. While these values are modest, they reinforce the interpretation that, while inconsistently, the model occasionally attempts to choose minority classes. 

The final observation is that the model consistently adhered to the schema, selecting only from the predefined set of valid values. Taken together with the preliminary results, this underscores both the feasibility of applying LLMs to CRF-filling and the need for systematic, large-scale experimentation to develop models that can meet clinical quality requirements.

\section{Conclusion}

We presented a case study on automatic CRF filling from emergency-care clinical notes. The central contribution of this work is the introduction of a new annotated dataset of 290 anonymised notes from the Emergency Department of San Giovanni Bosco Hospital in Turin, with annotations aligned to a 134-item CRF schema. This dataset captures the heterogeneity, sparsity, and imbalance of real-world documentation and is intended to provide a challenging yet realistic benchmark for future research on clinical information extraction.

To illustrate feasibility, we reported a pilot experiment with LLaMA-3 8B Instruct in a zero-shot setting. Preliminary results show that the model adheres to the schema and can occasionally retrieve non-trivial values, but out-of-the-box tools are not sufficient. There is a need for further work to engineer dedicated LLM-based solutions that can reach the quality required in clinical settings. These findings reinforce the importance of systematic and large-scale experimentation, building on the framework outlined in this paper.

At the time of publication, the dataset described here will be released to the research community, providing a shared resource for developing and evaluating models for CRF filling. We expect this release to enable reproducible experimentation and to accelerate progress on automatic structuring of clinical narratives in emergency-care settings.


\nocite{*}
\section{Bibliographical References}\label{sec:reference}

\bibliographystyle{lrec-coling2024-natbib}
\bibliography{lrec-coling2024-example}

\bibliographystylelanguageresource{lrec-coling2024-natbib}
\bibliographylanguageresource{languageresource}

\end{document}